%% file: example_paper.tex
\documentclass{article}

\usepackage{microtype}
\usepackage{graphicx}
\usepackage{subfigure}
\usepackage{booktabs} 

\usepackage{hyperref}

\usepackage{xcolor}
\usepackage{caption}
\usepackage{amsmath}
\usepackage{tabularx}
\usepackage{dblfloatfix}
\usepackage{array}
\usepackage{ragged2e}
\usepackage{xurl}
\usepackage[accepted]{mlsys2025}

\mlsystitlerunning{SparseLift: Automatically Optimizing Opaque Sparse Paths in Industrial-Scale Recommendation Model Training}

\begin{document}

\twocolumn[
\mlsystitle{SparseLift: Automatically Optimizing Opaque Sparse Paths in Industrial-Scale Recommendation Model Training}



\mlsyssetsymbol{equal}{*}

\begin{mlsysauthorlist}
\mlsysauthor{Changqing Shi}{tc}
\mlsysauthor{Anan Liu}{tc}
\mlsysauthor{Genbao Chen}{tc}
\mlsysauthor{Junchao Yang}{tc}
\end{mlsysauthorlist}

\mlsysaffiliation{tc}{Tencent, Beijing, China}

\mlsyscorrespondingauthor{Anan Liu}{ananliu@tencent.com}

\mlsyskeywords{Computation Graph; Deep Learning Compilers; Distributed Training; Large-Scale Recommendation Models; Machine Learning Systems; Sparse Computation}

\vskip 0.3in

\begin{abstract}
The sparse path, comprising sparse operations and collective communication, is a critical component of large-scale recommendation models. However, these operations are often represented as opaque nodes in the training computation graph, hindering whole-graph optimization. We present SparseLift, a low-overhead compiler for the sparse path of large-scale recommendation training. It exposes sparse computation and communication to whole-graph analysis to expand the optimization space and improve training efficiency. By analyzing model structure and runtime information, SparseLift reconstructs opaque nodes as analyzable subgraphs, recovering the resource semantics of sparse computation and communication within the graph. Building on the recovered semantics, we design incremental cost models and a constraint-aware planner to automatically select and organize subgraph optimizations into an execution plan that is reused during subsequent steady-state training. We evaluate SparseLift on an industrial video-platform recommendation task with 100B sparse parameters and the public DLRM benchmark. SparseLift improves steady-state throughput for ordinary industrial training steps by 16.5\% over the native baseline. On DLRM, it achieves $1.60\times$ and $1.26\times$ model-execution speedups over DLRM-Baseline and TorchRec, respectively.
\end{abstract}
]



\printAffiliationsAndNotice{}  

\section{Introduction}\label{0-Intra}
    \input{tex/0-Intra}
\section{Background and Motivation}\label{1-Back-Moti}
    \input{tex/1-Back-Moti}
\section{Design}\label{Design}
    \input{tex/2-Design}
\section{Evaluation}\label{Eval}
    \input{tex/3-Eval}
\section{Related Work}\label{Related}
    \input{tex/4-Related}
\section{Conclusion}\label{Conclu}
    \input{tex/5-Conclu}

\bibliography{example_paper}
\bibliographystyle{mlsys2025}

%


\section{AI Usage Disclosure}

During the research and writing of this paper, we used AI-assisted tools for writing refinement, assistance with code testing, and deployment guidance. These tools were used in a supporting capacity, and the paper’s content, code implementation, experimental results, and related materials were all handled and verified by us through hands-on work.

\clearpage

\appendix
\section{Decision Rules and Calibration Details}
\label{app:decision-details}
\input{tex/6-Decision-Details}

\end{document}

%% file: tex/0-Intra.tex
Training large-scale recommendation models has become one of the most important machine learning workloads in modern Internet services, while the continued growth of data and model sizes poses formidable challenges to training systems~\cite{Acun2020UnderstandingTE,Lian2021PersiaAO,Mudigere2021SoftwarehardwareCF}. In the industrial setting studied in this work, a single production use case generates billions of new training samples per day; the model contains hundreds of billions of sparse embedding parameters, as well as billions of dense network parameters. The massive sparse parameter space and its input-dependent, dynamic access patterns~\cite{Wang2024OPEROE} typically require the sparse path to span multiple devices, involving substantial per-table computation, input-dependent data reorganization, and frequent collective communication~\cite{Naumov2019DeepLR,Sethi2022RecShardSF,Zeng2024AcceleratingNR}. Its execution efficiency therefore directly affects training throughput, hardware utilization, and the efficiency of model updates~\cite{Liu2022MonolithRT}.

Prior work has explored sparse-path optimization along three main directions: parameter sharding~\cite{Sethi2022RecShardSF,Zha2022DreamShardGE,Zha2023PretrainAS}, specialized training systems~\cite{Zhang2022PICASSOUT,Liu2024EmbeddingOF,Huang2026FusedRecFE}, and domain-specific compilers~\cite{Pan2023REComAC,Pan2024RecFlexEF}. However, these approaches have yet to provide a unified analysis of the sparse path's internal structure, stage boundaries, resource semantics, and cross-stage and cross-iteration execution relationships within the training computation graph to support whole-graph rewriting and scheduling. Moreover, they are often tightly coupled to specific runtimes and rely on predefined rules or fixed compatibility conditions, limiting their ability to automatically select and compose optimization strategies as training tasks evolve.

Collectively, these studies reveal substantial optimization opportunities along the sparse path, while also showing that existing capabilities remain fragmented across disconnected layers of parameter planning, runtime mechanisms, and local compilation~\cite{Lattner2021MLIRSC,Bik2022CompilerSF,10.1145/3503222.3507778,Ansel2024PyTorch2F,Jia2019TASOOD,Wang2024RAPRA}. Recovering the internal semantics of opaque sparse path, making sparse operations visible and analyzable entities that can participate in compiler-directed decisions, and thereby expanding the whole-graph optimization space therefore remain essential to improving the efficiency of large-scale recommendation model training.

Our key insight is that incorporating the sparse path into whole-graph optimization does not require reconstructing the runtime in its entirety. Instead, lifting the minimal semantic and resource facts that determine optimization opportunities, legality, and profitability is sufficient to reconstruct transparent subgraphs that support compiler optimization. Furthermore, the structural semantics of these subgraphs can define candidate transformations and their feasible space, while their incremental costs relative to the incumbent execution plan provide reliable evidence for strategy selection.

Building on these insights, we present SparseLift, a low-overhead compiler for the sparse path of large-scale recommendation training. SparseLift recovers fine-grained sparse operations and their execution relationships along semantic boundaries to construct analyzable subgraphs. We develop incremental cost models for subgraph optimizations and design a constraint-aware planner to automatically select and organize them into execution plans. We evaluate SparseLift on a production training task and the public DLRM benchmark. On the industrial task, SparseLift improves steady-state throughput for ordinary training steps by 16.5\% over the native baseline. On DLRM, it achieves up to a $1.26\times$ model-execution speedup over TorchRec. Our key contributions are as follows:

\begin{itemize}

\item We propose sparse semantic lifting, which analyzes model structure and runtime information to reconstruct key semantics and resource facts as analyzable subgraphs amenable to whole-graph optimization, addressing the structural mismatch between the critical sparse path and its semantic opacity in the computation graph.

\item We design incremental cost models and decision algorithms that define the feasible space using the recovered semantics and execution constraints. They automatically select subgraph optimization strategies based on cost changes relative to the incumbent plan. This enables more accurate assessment of optimization benefits despite the various sources of interference encountered during production training.

\item We design a constraint-aware planner to automatically organize the selected subgraph optimization strategies into graph transformations and scheduling decisions. We validate SparseLift's improvements in production training efficiency on a video recommendation platform serving approximately one billion users.

\end{itemize}

%% file: tex/1-Back-Moti.tex
\subsection{Sparse-Path Structure and Optimization Opportunities}
\label{sec:sparse-opportunities}

\begin{figure}[!h]
  \centering
  \includegraphics[width=0.75\columnwidth]{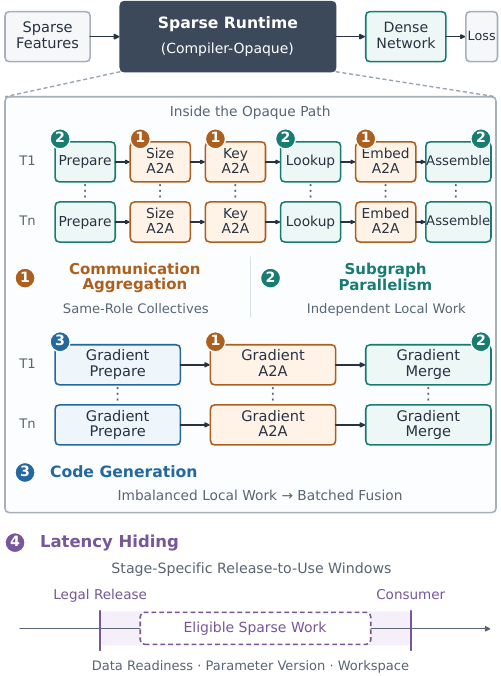}
  \caption{Schematic of the sparse path and optimization opportunities in the industrial video-platform task. A2A: all-to-all; T: embedding table.}
  \label{fig:hidden-sparse-path}
\end{figure}

The sparse path in recommendation training connects per-table computation, parameter access, and cross-device communication~\cite{Naumov2019DeepLR,Mudigere2021SoftwarehardwareCF}. In the industrial video-platform task studied in this work (Figure~\ref{fig:hidden-sparse-path}), the forward pass prepares and routes inputs, looks up parameters, and assembles embeddings, while the backward pass prepares, exchanges, and merges gradients. These per-table repetitions and stage boundaries may be hidden behind opaque runtime calls. Optimization opportunities therefore depend not only on individual-operator efficiency but also on how per-table work is organized and coordinated with other training stages~\cite{10.1145/3503222.3507778}.

These execution relations offer four classes of optimization opportunities. Compatible collectives with matching roles within a stage can be aggregated to amortize call overhead~\cite{Huang2026FusedRecFE,Sergeev2018HorovodFA}; independent per-table computations can execute concurrently while retaining member implementations~\cite{Ma2020RammerEH}; and structurally compatible member computations can be jointly compiled to reduce repeated launches and intermediate processing~\cite{Pan2023REComAC,Pan2024RecFlexEF,Li2020AutomaticHF}. When a legal window exists between readiness and result consumption, sparse work can execute early and overlap other stages, subject to constraints such as parameter versions and buffer lifetimes~\cite{Jayarajan2019PrioritybasedPP,Peng2019AGC}. The profitability and compatibility of these opportunities depend on workloads, resource conditions, and existing execution arrangements, motivating task-specific selection and organization of optimization strategies.

\begin{figure*}[!t]
  \centering
  \includegraphics[width=0.83\textwidth]{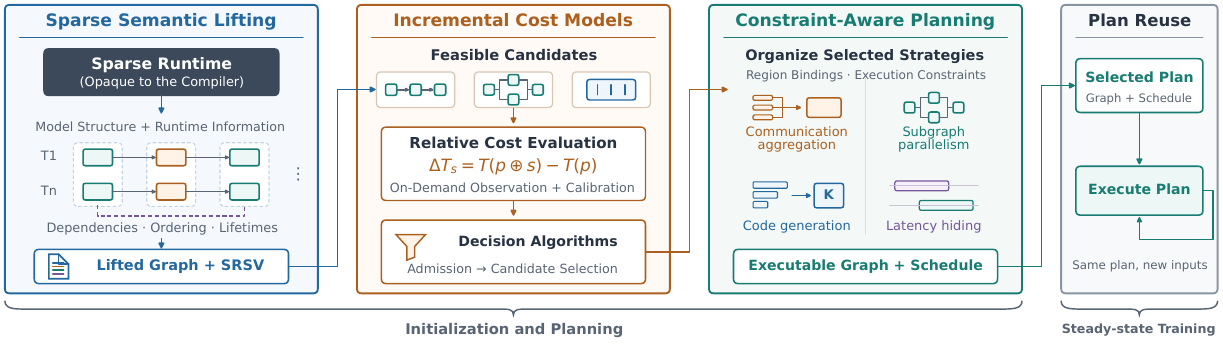}
    \caption{SparseLift architecture: sparse semantic lifting, incremental cost-based decisions, and constraint-aware planning. The resulting execution plan is reused during steady-state training. SRSV: Sparse Resource Semantic View; $T_i$: Embedding table $i$.}
  \label{fig:sparselift-architecture}
\end{figure*}

\subsection{Challenges in Automatic Sparse-Path Optimization}

Automatically optimizing the sparse path in industrial training requires recovering sufficient training semantics for graph-level analysis while preserving runtime encapsulation. Exposing only call inputs and outputs does not distinguish cross-table independence from the baseline's serial execution order, whereas expanding the entire runtime introduces implementation details irrelevant to optimization. Beyond forward data dependencies, forward--backward correspondences, collective ordering, parameter versions, and lifetimes spanning batches jointly constrain legal transformations; local numerical equivalence alone does not permit arbitrary reordering~\cite{Narayanan2019PipeDreamGP,10.1145/3696630.3728529}. Per-member workloads and communication volumes also affect candidate profitability~\cite{Wang2024OPEROE}. The challenge is therefore to recover information sufficient for opportunity identification, legality checks, and cost assessment while retaining existing runtime mechanisms.

Optimization benefits depend on the execution context in which a strategy operates, making candidate selection inseparable from how strategies are executed together. Aggregation, parallelism, and generated implementations change communication granularity, resource contention, and region execution times, which in turn affect latency-hiding benefits; gains measured in isolation therefore cannot simply be added. In industrial training environments, asynchronous waiting and platform interference may also obscure candidate-induced timing changes, making direct comparisons of observed total execution times insufficient to distinguish strategy effects from environmental variation~\cite{Hashemi2018TicTacAD,Mytkowicz2009ProducingWD}. Decisions must therefore rely on comparable evidence of relative cost in the context established by earlier choices, while organizing the selected strategies requires coordinating region implementations, execution order, and resource lifetimes so that local choices yield graph transformations and schedules that satisfy training execution constraints.

%% file: tex/2-Design.tex
\subsection{Overview}
\label{sec:design-overview}
As shown in Figure~\ref{fig:sparselift-architecture}, SparseLift organizes sparse-path compilation and optimization into three interconnected components. First, sparse semantic lifting analyzes model structure and runtime information to reconstruct opaque nodes as analyzable subgraphs and a read-only Sparse Resource Semantic View (SRSV), exposing structural relations, execution constraints, and information for cost queries. Second, incremental cost models and decision algorithms evaluate feasible candidates by their cost changes relative to the current plan, automatically selecting applicable candidates for communication aggregation, subgraph parallelism, code generation, and latency hiding. Finally, the constraint-aware planner binds the selected candidates to their corresponding sparse regions and execution points, organizing graph transformations and scheduling while preserving training execution constraints to enable cross-table, cross-stage, and cross-iteration optimization. This process takes place during initialization and planning; the resulting execution plan is reused throughout steady-state training for the same task while its applicability conditions hold, avoiding replanning at every step.

\subsection{Sparse Semantic Lifting}
\label{sec:semantic-lifting}

Sparse semantic lifting establishes an optimization-oriented semantic representation that distinguishes training execution relations that must be preserved from implementation choices that can be reorganized. To this end, we recover operation and resource facts according to optimization requirements, use execution constraints and transformation contracts to identify transformable sparse regions and legal candidates, and expose consistent information for subsequent decisions through the read-only SRSV. This representation grounds optimization opportunities, feasibility conditions, and cost information in a consistent set of semantic objects, enabling the recovered sparse structure to directly support strategy selection and organization.

\begin{table*}[!h]
  \centering
  \caption{Mapping recovered semantics to optimization feasibility and cost-based decisions.}
  \label{tab:semantic-optimization-map}
  \small
  \setlength{\tabcolsep}{4pt}
  \renewcommand{\arraystretch}{1.03}
  \begin{tabularx}{\textwidth}{@{}l*{3}{>{\RaggedRight\arraybackslash}X}@{}}
    \toprule
    Optimization
    & Recovered relation
    & Feasibility checks
    & Cost inputs and decision \\
    \midrule

    Communication aggregation
    & Collectives with matching semantic roles
    & Communication group; data type; layout mapping; collective ordering
    & \textit{Inputs:} per-member traffic; packing/unpacking cost.
      \textit{Choice:} roles to aggregate. \\

    Subgraph parallelism
    & Independent table-local operations
    & Dependencies; workspace requirements; stream concurrency support
    & \textit{Inputs:} per-member workloads; orchestration overhead.
      \textit{Choice:} group count $G$. \\

    Code generation
    & Structurally compatible repeated computation
    & Value/gradient correspondence; address isolation; interface compatibility
    & \textit{Inputs:} workload distribution; region execution time.
      \textit{Choice:} implementation $v$. \\

    Latency hiding
    & Release-to-use relations
    & Input readiness; parameter version; batch identity; buffer lifetime
    & \textit{Input:} complete schedule execution time.
      \textit{Choice:} enable/disable. \\

    \bottomrule
  \end{tabularx}
\end{table*}

\textbf{Optimization-Directed Semantic Recovery.} We bound semantic recovery by the information requirements of candidate transformations, using structural relations, execution constraints, and workload features to support opportunity identification, legality checks, and cost queries, respectively. Specifically, adapters export operation facts and typed ports from module metadata, runtime interfaces, and lifecycle contracts to construct the sparse semantic graph $G_R$, while on-demand observation supplies dynamic workload information. Here, $N_S$, $N_C$, and $N_{BW}$ represent forward local operations, collectives, and backward operations, respectively. To capture training execution conditions beyond dataflow, we augment data and metadata dependencies $E_D$ with execution constraints $E_X$ that encode parameter versions, buffer ownership, collective ordering, and batch identity. This representation gives the compiler the training semantics needed to analyze candidate transformations without exposing every runtime implementation detail.
\[
G_R=\left(N_S\cup N_C\cup N_{BW},\;E_D\cup E_X\right).
\]

\textbf{Transformable Structure and Legality.} To avoid mistaking runtime call order for semantic dependencies, we distinguish execution relations that must be preserved from implementation orderings that may be changed: per-table calls do not inherently imply dependencies, whereas parameter-version readiness and resource release constrain legal execution points. Accordingly, the compiler partitions operations at communication-domain, parameter-version, resource-ownership, and consumption boundaries, then combines compatible operations into sparse regions through declarative patterns. The forward graph and a backward semantic overlay connect these regions to dense consumers, Autograd, and the optimizer lifecycle. We combine recovered dependency, ordering, and lifetime facts with the value correspondences and available implementations specified by transformation contracts to construct the feasibility predicate $\Phi_s$. This predicate allows candidates to reorganize execution while preserving data semantics and applicable constraints under their input--output mappings.

\begin{figure}[!t]
  \centering
  \includegraphics[width=0.91\columnwidth]{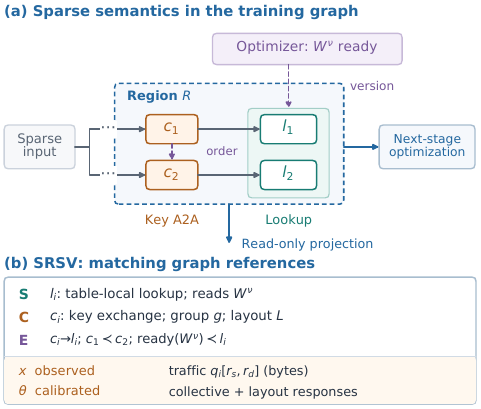}
  \caption{Lifted graph slice and read-only SRSV. Gray: dataflow; purple: execution constraints; blue: compiler projection or handoff. A2A: all-to-all.}
  \label{fig:sparse-semantic-lifting}
\end{figure}

\textbf{From Semantics to Decisions.} To ground the analysis of different strategies in consistent semantic objects, we separate the transformable graph from the read-only SRSV, which supplies decision facts through stable region and member references (Figure~\ref{fig:sparse-semantic-lifting}). These facts inform feasibility conditions and strategy-specific cost queries, providing a shared semantic basis for enforcing execution constraints and assessing performance benefits. To limit evidence-collection overhead, we scope observation to candidate requirements: retaining per-member workloads and source--destination rank traffic, checking capacity for additional buffers, and performing targeted calibration at safe points when timing measurements are missing. Evidence is bound to implementation variants, graphs, and environments, preserving workload differences across members while delimiting the conditions under which measurements can be reused. Table~\ref{tab:semantic-optimization-map} shows how this information supports the four optimizations.

\subsection{Incremental Cost Models}
\label{sec:incremental-planning}

Incremental cost models assess candidate benefits in the execution context of the current plan, accounting for both interactions among optimizations and measurement interference in industrial training. Using relative cost changes as a common basis, we combine strategy-specific cost tradeoffs with bounded calibration and shared procedures for evidence processing and applicability checks to establish admission and selection rules for communication aggregation, subgraph parallelism, code generation, and latency hiding. The selected candidates are then passed to the constraint-aware planner for organization into graph transformations and scheduling; when evidence is insufficient or inapplicable, the current choice is retained.

\textbf{Context-Dependent Incremental Decisions.} To account for the effects of previously selected strategies, we define the increment $\Delta T_s=T(p\oplus s)-T(p)$ relative to the validated incumbent plan $p$, where $p\oplus s$ denotes the plan after applying the candidate, and $T$ measures the execution time of the affected region or complete schedule. We combine analytical cost tradeoffs with timing calibration in the target environment to select feasible candidates based on relative cost. Algorithm~\ref{alg:sequential-planning} evaluates communication aggregation, subgraph parallelism, and code generation in sequence, updating the context for subsequent evaluations with earlier choices before assessing latency hiding under the established region bindings. The algorithm takes the SRSV $V$, transformation contracts $\mathcal C$, initial plan $p_0$, and budget and policy parameters $\Theta$ as inputs; each decision unit $u$ is a region, a set of compatible communication roles, or a complete schedule. We use $\oplus_{\mathcal C}$ to merge compatible bindings while maintaining implementation exclusivity and cross-strategy constraints and retaining all other incumbent bindings.

\begin{algorithm}[!t]
\caption{Incremental Strategy Selection}
\label{alg:sequential-planning}
\small
\renewcommand{\algorithmiccomment}[1]{\hspace{0.5em}\textnormal{// #1}}
\begin{algorithmic}[1]
\STATE {\bfseries Input:} view $V$, contracts $\mathcal C$,
       validated plan $p_0$, policy $\Theta$
\STATE $p\gets p_0$;
       $\mathcal S\gets(\mathrm{agg},\mathrm{par},\mathrm{cg},\mathrm{LH})$
\FORALL{$s\in\mathcal S$ in order}
  \STATE $b\gets p$; $\mathcal B\gets\emptyset$
         \COMMENT{Fix reference}
  \STATE $\mathcal P\gets
         \operatorname{EnumerateCandidates}_s(V,\mathcal C,b)$
  \STATE $\mathcal F\gets
         \operatorname{FilterFeasible}_s(\mathcal P;V,\mathcal C,b)$
  \IF{$\mathcal F\ne\emptyset$}
    \STATE $Q\gets
           \operatorname{CollectQueries}_s(V,\mathcal F,b)$
    \STATE $D\gets
           \operatorname{Calibrate}_s(Q;b,\Theta_s)$
           \COMMENT{Bounded trials}
    \FORALL{$u\in\operatorname{DecisionUnits}_s(\mathcal F,D)$}
      \IF{$\operatorname{ValidEvidence}(D|_u,V|_u,b)$}
        \STATE $E_u\gets
               \operatorname{BuildEvidence}_s(D|_u,Q|_u)$
        \STATE $\nu_u\gets
               \operatorname{SelectCandidate}_s(E_u;\Theta_s)$
               \COMMENT{Rules below}
        \STATE $\mathcal B\gets
               \operatorname{BindChoice}(\mathcal B,u,\nu_u)$
      \ENDIF
    \ENDFOR
    \STATE $p\gets
           \operatorname{MergeBindings}(b,\mathcal B;\mathcal C)$
  \ENDIF
\ENDFOR
\STATE {\bfseries return} $\operatorname{FreezePlan}(p)$
       \COMMENT{Reuse in training}
\end{algorithmic}
\end{algorithm}

\textbf{Shared Evidence and Applicability Boundaries.} To ensure that calibration results apply to final execution, we perform bounded measurements after warmup and keep the measured implementations consistent with the plan bindings. For timing variability dominated by added waiting, we use the lower-tail statistic $F(X)=Q_q^r(\min_jX_{r,j})$, taking the minimum over samples within each rank and then the $q$-quantile across ranks. Communication aggregation, subgraph parallelism, and code generation use paired increments $d_s=F(\delta_s)$, where $\delta_s$ is the execution-time difference between a candidate and its corresponding reference, to mitigate shared environmental variation. Subgraph parallelism and code generation additionally interpolate neighboring reference measurements to mitigate drift. Latency hiding instead compares lower-tail execution times of complete schedules to reduce the influence of added waiting. Because residual interference in paired differences can have either sign, we use $F(\delta_s)$ as an empirical decision statistic rather than an estimator with statistical confidence guarantees.

To make decisions under limited measurement resolution, we combine structural opportunities with measured evidence to define strategy-specific admission rules. We define $(x)_+ = \max(0,x)$ and use $\mathcal{M}_s(f,u) = \max\{\epsilon_s,\tau_s f,\kappa_s u\}$ to combine absolute resolution, relative response scale, and an interference budget. In the following, $\mathbf{1}[\cdot]$ denotes the indicator function; $a^*$, $c^*$, and $\ell^*$ equal 1 when communication aggregation, the generated implementation, and latency hiding are selected, respectively. When evidence is missing, invalid, or inapplicable, we retain the incumbent binding; otherwise, we make selections according to the strategy-specific rules.

\textbf{Communication Aggregation.} We compare the cost of per-member communication with that of the complete pack--collective--unpack region to weigh fewer calls against the added overhead. $Q=\{q_i\}$ records inter-rank traffic for $N$ compatible members, and $q_\Sigma=\bigoplus_i q_i$ aggregates matching source--destination entries; $\widehat L_{pu}$ denotes packing and unpacking costs, including layout conversion, and $\widehat C$ denotes communication time. Define the reference response $f_0=F(T_0)$, interference budget $U=(\operatorname{med}_j\max_r T_{0,r,j}-f_0)_+$, and call-reduction fraction $P=1-1/N$:
\begin{gather*}
\Delta\widehat T_{\mathrm{agg}}=\widehat L_{pu}(Q)+\widehat C(q_\Sigma)-\sum_i\widehat C(q_i),\displaybreak[1]\\
\varepsilon_{\mathrm{agg}}=\max\{\epsilon_a,\tau_af_0,U,\lambda_aPf_0\},\displaybreak[1]\\
a^*=\mathbf1[d_{\mathrm{agg}}\le\varepsilon_{\mathrm{agg}}].
\end{gather*}
We use $P$ to characterize the structural scale of launch amortization and paired measurements of the complete region to reject regressions beyond tolerance. Joint forward candidates are measured over the entire chain; when joint evidence is unavailable, decisions are made per communication role. Backward communication is evaluated independently.

\textbf{Subgraph Parallelism.} To balance parallelism gains against orchestration and contention overheads, we model the cost of region $h$ over executor-provided feasible group counts $\mathcal G$ and fixed partitions $\pi_G$, with serial execution within groups and concurrency across groups. The model combines calibrated responses $\widehat f_h(\xi_i)$ for member workload features $\xi_i$, orchestration overhead $H(G)$, and a contention factor $\rho_h(G)$. Let $f_G=F(T_G)$ and $d_G=F(\delta_G)$, and define $M_G=\operatorname{med}_jQ_q^r(T_{G,r,j})$ as the median across samples of the cross-rank quantile responses. We use the structural score $S_G=1-\max_g|\pi_G(g)|/N-H(G)/M_1$ to assess member grouping against orchestration cost, while region measurements capture workload differences and actual contention; $N$ is the total number of members. The selection rule is:
\begin{gather*}
\widehat T_h(G)=H(G)+\rho_h(G)\max_g\sum_{i\in\pi_G(g)}\widehat f_h(\xi_i),\displaybreak[1]\\
\varepsilon_G=\max\{\epsilon_p,\tau_pf_1,U_0,U_G\}+\kappa_vH^{cal},\displaybreak[1]\\
\mathcal A=\{G\in\mathcal G:G>1,\ S_G>0,\ d_G\le\varepsilon_G\},\displaybreak[1]\\
G^*=\begin{cases}
1,&\mathcal A=\emptyset,\\
\min\{G\in\mathcal A:d_G\le d_{\min}+\omega\},&\text{otherwise}.
\end{cases}
\end{gather*}
We use $U_0=(M_1-f_1)_+$ and $U_G=\min\{(M_G-f_G)_+,f_1/G\}$ to limit noise-based tolerance relaxation, and define $d_{\min}=\min_{G\in\mathcal A}d_G$ and $\omega=\max\{\tau_p f_1,\kappa_w H^{cal}\}$, where $H^{cal}$ is the orchestration timing resolution. Among the given partitions, we prefer configurations with fewer groups within the equivalence band to avoid unnecessary resource overhead.

\textbf{Code Generation.} To determine whether fewer launches from reorganizing computations across members offset the auxiliary overhead, we compare the generated implementation of a complete region $\mathcal R$ with the selected nongenerated reference, modeling execution time $\widehat T_{\mathcal R}(v)$ as a function of member workloads and the target environment. Let $d_{\mathrm{cg}}=F(\delta_{\mathrm{cg}})$. Within each rank, we compute the sample ranges of reference times, generated times, and paired differences, then define $\eta$ as the cross-rank quantile of the maximum of these three ranges and set $m_{\mathrm{cg}}=\mathcal M_{\mathrm{cg}}(F(T_{\mathrm{ref}}),\eta)$:
\begin{gather*}
\Delta\widehat T_{\mathrm{cg}}=\widehat T_{\mathcal R}(v_{\mathrm{gen}})-\widehat T_{\mathcal R}(v_{\mathrm{ref}}),\displaybreak[1]\\
c^*=\mathbf1\!\left[d_{\mathrm{cg}}<0\ \lor\
 (d_{\mathrm{cg}}\le m_{\mathrm{cg}}\land B_{\mathcal R})\right].
\end{gather*}
We select the generated implementation when the measured increment is negative. A nonnegative increment within tolerance is admitted only when the member count and workload skew captured by $B_{\mathcal R}$ support reorganizing work across members; the specific rule appears in Appendix~\ref{app:decision-details}. To assess the implementation's overall benefit, we measure the complete region, including metadata processing, sorting, output projection, and kernel execution.

\textbf{Latency Hiding.} Within the recovered legal overlap windows and under the current region bindings, we compare complete step times with one-step lookahead enabled and disabled to assess the overall benefit. Enabled samples are included only when the intended schedule executes and the prefetched results are consumed. With $a=0/1$ denoting disabled/enabled execution, define $f_{a,r}=\min_j T_{a,r,j}$, $d_{\mathrm{LH}}=F(T_1)-F(T_0)$, and $U_r=\sum_{a\in\{0,1\}}(\operatorname{med}_j T_{a,r,j}-f_{a,r})_+$, and let $m=\mathcal M_{\mathrm{LH}}(F(T_0),Q_q^r(U_r))$ and $m_r=\mathcal M_{\mathrm{LH}}(f_{0,r},U_r)$:
\begin{gather*}
v_{\mathrm{LH}}=[d_{\mathrm{LH}}>m]\land\bigwedge_r[f_{1,r}-f_{0,r}>m_r],\displaybreak[1]\\
\ell^*=\mathbf1[\neg v_{\mathrm{LH}}].
\end{gather*}
To prevent local anomalies from dominating the overall decision, we veto a candidate only when both the cross-rank summary and all local responses exceed their respective regression thresholds. All four rules require candidates to pass legality and evidence-validity checks and use references constrained by contracts; policy parameters appear in Appendix~\ref{app:decision-details}.

\begin{figure}[!h]
  \centering
  \includegraphics[width=0.91\columnwidth]{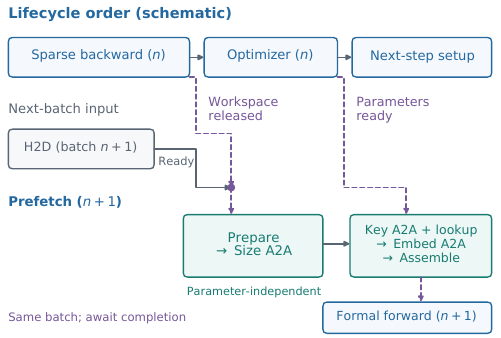}
  \caption{Latency hiding and its execution constraints. Dashed edges denote workspace availability, parameter-version, and result-consumption constraints.}
  \label{fig:latency-hiding-contract}
\end{figure}

\subsection{Constraint-Aware Planning}
\label{sec:graph-realization}

The constraint-aware planner binds selected candidates to their target regions, implementations, and execution points, organizing them into compatible graph transformations and schedules that translate local strategy choices into coordinated cross-table, cross-stage, and cross-iteration optimization. Guided by recovered execution constraints and implementation contracts, we coordinate the combination and execution of subgraph optimizations, using cross-rank consistency checks and runtime dependency management to preserve training semantics throughout their integration and execution.

We realize the selected strategies as graph transformations on semantic regions, guided by value correspondences and execution constraints. For communication aggregation, we rewrite compatible calls with matching semantic roles into pack--collective--unpack subgraphs, preserving member input--output mappings through source/destination layouts and element counts while updating collective sequences consistently across ranks. For subgraph parallelism, we retain implementations of independent members that support concurrency, execute serially within groups and concurrently across groups, and use GPU events and ownership management to preserve cross-stream dependencies and tensor lifetimes. Code generation instead replaces region implementations with batched implementations that reorganize structurally compatible computations within regions and across members while preserving value and gradient correspondences, address isolation, and the specified numerical semantics.

We use the recovered readiness and lifetime constraints to determine early-execution boundaries (Figure~\ref{fig:latency-hiding-contract}): parameter-independent preparation executes early once inputs are ready and workspace is released, while parameter access waits for the required version. We use batch-identity checks and GPU events to ensure that consumers receive matching, completed prefetch results, and maintain consistent communication groups, call counts, and ordering across ranks on both the prefetch-hit and prefetch-miss paths of the one-step-lookahead schedule. Sparse work can thus overlap other stages within legal windows, reducing its exposed time on the training critical path.

Before activation, we require each rank to check the graph, environment, plan, evidence, and communication-resource identifiers; if a check fails, all ranks retain the validated plan. If a failure occurs after partial submission, we terminate execution after synchronization and resource cleanup to prevent replay from causing duplicate communication or writes.

\subsection{Implementation}
\label{sec:implementation}

We integrate SparseLift with PyTorch model classes or sparse modules through \texttt{sparselift.compile}. An adapter written once per runtime enables automatic semantic lifting, decision-making, planning, and graph transformation for each training task without task-specific code. We perform analysis and planning on the domain-specific semantic graph constructed by the adapter, using module execution interfaces and training lifecycle hooks to connect the forward pass, Autograd, and the optimizer. To expose cross-table backward relations, we introduce batched Autograd boundaries after capability and correctness checks and establish Autograd connections on the main thread. Selected transformations execute through the runtime's communication, stream, and event interfaces, with generated kernels implemented in Triton, while preserving the original parameter-access interfaces and parameter-server storage mechanism.

We compile and tune the generated code before collecting timing samples and share the resulting artifacts across ranks. Reusing the SRSV, queries, and decision logic across runtimes requires adapting fact export, lifecycle integration, and candidate execution interfaces, along with recalibration in the target environment. We treat host-thread and input-queue coordination as supporting infrastructure, decoupled from graph-plan selection.

%% file: tex/3-Eval.tex
\subsection{Experimental Setup}

\textbf{Platform and Workloads.} Table~\ref{tab:experimental-setup} lists the configurations of our experimental environment. We conduct the industrial evaluation within a video platform's continuous training pipeline, using training data derived from real-world statistics collected from over one billion users of the platform. The model contains 1B dense parameters and 100B sparse parameters. It constructs input representations through multi-table embeddings, attention over behavioral sequences, and feature interactions, followed by a RankMixer~\cite{10.1145/3746252.3761507} backbone and multitask prediction towers. The industrial experiments cover 1, 8, and 16 GPUs, with the 16-GPU configuration spanning two identically configured nodes. All industrial experiments use a batch size of 2048, with sparse parameters stored on dedicated parameter servers. The public DLRM~\cite{Naumov2019DeepLR} benchmark uses the Criteo Kaggle dataset~\cite{criteo2014kaggle} and runs on 1/2/4/8 GPUs within a single node, with a global batch size of 16384, an embedding dimension of 256, FP32 precision, and SGD with a learning rate of 0.1.

\begin{table}[!b]
  \centering
  \caption{Experimental platform specifications per node.}
  \label{tab:experimental-setup}
  \small
  \setlength{\tabcolsep}{4pt}
  \begin{tabularx}{\columnwidth}{@{}lX@{}}
    \toprule
    Component & Configuration \\
    \midrule
    GPUs & $8\times$ H20 (96 GiB each) \\
    CPUs & $2\times$ EPYC 9K84 (96 cores each) \\
    Host memory & 2.2 TiB \\
    Intra-node interconnect & NVLink \\
    Network port rate & 200 Gb/s Ethernet \\
    \midrule
    PyTorch / CUDA & 2.6.0 / 12.6 \\
    NCCL & 2.21.5 \\
    TorchRec / FBGEMM-GPU & 1.1.0 / 1.1.0 \\
    Triton & 3.2.0 \\
    \bottomrule
  \end{tabularx}
\end{table}

\textbf{Baselines and Comparisons.} The industrial evaluation uses the native training implementation as the baseline. Its dense path has already been optimized through manual tuning and \texttt{torch.compile}, while its sparse path follows the production runtime's existing execution scheme. Given the industrial task's integration with dedicated parameter servers and the training pipeline, we preserve these execution conditions to assess the benefits of SparseLift's sparse-path optimizations within the existing system. We denote the complete optimization plan automatically selected and organized by SparseLift as Full and disable individual strategies to examine their contributions.

Beyond the industrial task, we use the public DLRM benchmark to assess SparseLift's performance under a different workload and include TorchRec~\cite{torchrec2022} as a cross-stack reference. DLRM-Baseline and SparseLift share the same model, data-processing pipeline, and training configuration, providing a controlled comparison of execution optimizations. TorchRec retains its default optimization configuration and native execution stack for DLRM, along with its own data preprocessing and data split.

\textbf{Measurement Methodology.} We compute industrial throughput from the primary training process's logs over the first 1000 steps, excluding the first 60 steps and the first step of each 10-step period. For each remaining period, we divide the total sample count by the total execution time of its nine included steps and report the median and 25th--75th percentile interval across 94 steady-state periods. Industrial throughput is measured with profiling enabled, and the stage breakdown uses four stage-timing records per configuration. Sparse forward and backward times include communication; the remaining forward, backward, and Other times are derived by subtracting constituent stages from their enclosing totals. Each DLRM configuration is run independently five times, with 50 warmup steps followed by 450 measured steps per run. We first take the median across the nine measurement windows within each run, then the median across the five runs. DLRM performance is measured using CUDA events with profiling disabled, excluding data loading and host-to-GPU transfer overhead. We assess training quality using loss over the first 1000 industrial training steps and test metrics after one complete training epoch for DLRM.

\subsection{Industrial Performance and Stage Breakdown}
\begin{figure}[!t]
  \centering
  \includegraphics[scale=0.8]{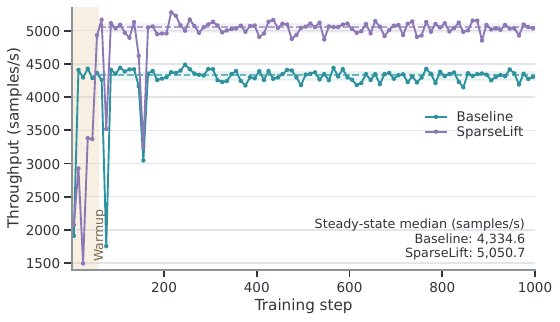}
  \caption{Industrial throughput on 8 GPUs. Points aggregate nine ordinary steps; bands/dashes show IQRs/medians after step 60.}
  \label{fig:industry-throughput}
\end{figure}

Figure~\ref{fig:industry-throughput} compares SparseLift with the native baseline on the industrial task. On 8 GPUs, median steady-state throughput for ordinary training steps increases from 4334.6 to 5050.7 samples/s, a gain of 16.5\%. The upward shift in the interquartile interval indicates that the gains extend across typical steady-state periods. Building on a baseline with an already optimized dense path, SparseLift recovers sparse execution relations to bring cross-table computation and communication into a unified analysis, reducing sparse-processing overhead and coordinating sparse work with other training stages to further improve training throughput.

The stage breakdown in Figure~\ref{fig:industry-ablation}(a) further reveals the sources of these gains. On 8 GPUs, sampled step time decreases from 451.5 to 378.5 ms, a reduction of 16.2\%. Sparse forward time decreases from 44.25 to 3.13 ms, while sparse backward time decreases from 28.00 to 9.13 ms. The substantial reduction in exposed forward time is consistent with latency hiding: recovered readiness and lifetime constraints establish legal windows for early execution, allowing some results to be ready before their consumers execute and thereby reducing waiting on the consumption path. The backward pass exploits recovered cross-table compatibility through communication aggregation and code generation to reduce communication startup and computation scheduling overhead. Together, these results reflect the combined effects of reducing sparse-processing overhead and its exposure on the training critical path. Sampled step times also decrease substantially relative to the native baseline on 1 and 16 GPUs, indicating that the optimization benefits extend to larger-scale execution of this training task.

\begin{table}[!t]
  \centering
  \caption{Workload-specific execution choices. Keep ref. denotes retaining the reference execution.}
  \label{tab:execution-choices}
  \small
  \begin{tabularx}{\columnwidth}{@{}lXl@{}}
    \toprule
    Strategy & Industry & DLRM \\
    \midrule
    Aggregation & size/key/embed/grad & Embeddings \\
    Parallelism & $2/2/1/4$ groups\textsuperscript{a} & Keep ref. \\
    Code generation & Gradient preparation & Keep ref. \\
    Latency hiding & One-step lookahead & Keep ref. \\
    \bottomrule
    \multicolumn{3}{@{}p{\columnwidth}@{}}{\footnotesize \textsuperscript{a}Input preparation / output assembly / gradient preparation / gradient merging. Generated gradient preparation requires one group.} \\
  \end{tabularx}
\end{table}

\subsection{Automatic Plan Selection and Strategy Contributions}

Table~\ref{tab:execution-choices} presents the execution plans automatically selected for the two workloads on 8 GPUs. The industrial plan identifies aggregation targets by communication role and selects parallel group counts and implementations for individual sparse regions. Gradient preparation uses a generated implementation whose contract requires single-group execution, while other regions retain member implementations and execute in groups. These choices illustrate how cross-table computation restructuring and concurrency organization are jointly selected according to region characteristics. Latency hiding further uses recovered execution constraints to identify legal windows. Its complete-schedule calibration yields 384.13 ms, compared with 418.75 ms for the incumbent reference, providing cost evidence for adopting one-step lookahead. Together, these choices translate recovered semantics into concrete region implementations and execution schedules that are reused during subsequent training.

\begin{figure*}[!t]
  \centering
  \includegraphics[scale=0.8]{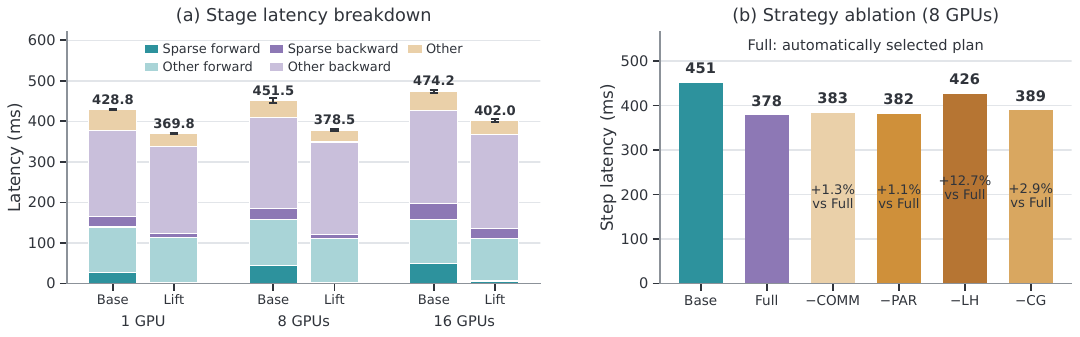}
  \caption{Industrial stage breakdown and ablation. (a) Base/Lift: Baseline/SparseLift; error bars: total-step standard deviations over four measurements. (b) Percentages show latency increases upon individual strategy removal from Full.}
  \label{fig:industry-ablation}
\end{figure*}

Figure~\ref{fig:industry-ablation}(b) examines each strategy's contribution within the complete plan by disabling it individually. Full has a sampled step time of 378 ms; disabling communication aggregation, subgraph parallelism, latency hiding, or code generation increases latency by 1.3\%, 1.1\%, 12.7\%, or 2.9\%, respectively. Latency hiding has the largest effect, consistent with both the direction of the calibrated benefit and the reduction in exposed forward time, indicating that legal overlap windows allow sparse work to execute early and reduce consumer waiting. The remaining strategies further reduce execution overhead through communication amortization, concurrent execution of independent work, and cross-table computation restructuring. Full performs best among these ablation configurations, supporting the joint contribution of region-level execution optimizations and cross-stage scheduling to industrial training performance.

For DLRM, cost evaluation selects embedding communication aggregation while retaining the reference execution for the other strategies. Per-table embedding exchanges offer a clear opportunity to amortize call overhead, whereas local sparse computation and scheduling opportunities are more limited, consistent with the different choices made for this workload. Subgraph parallelism requires sufficient independent work to amortize orchestration overhead, generated implementations require gains from cross-member restructuring to offset auxiliary processing costs, and latency hiding must balance overlap benefits against scheduling overhead within legal windows. These conditions make strategy applicability dependent on the specific region and execution context. The differences between the two plans show how SparseLift combines recovered structure with cost evidence to make selections while allowing each workload to retain execution choices suited to its characteristics.

\subsection{DLRM Performance}
\begin{figure*}[!t]
  \centering
  \includegraphics[scale=0.75]{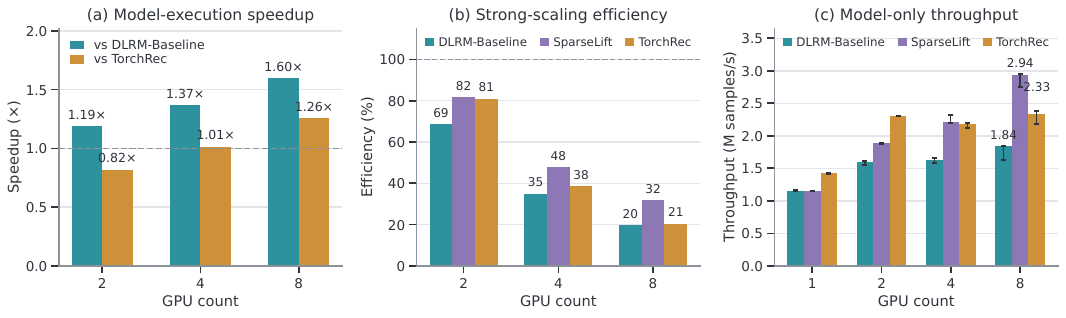}
  \caption{DLRM model-execution performance, reported as five-run medians. (a) SparseLift speedups over DLRM-Baseline and TorchRec. (b) Each system's single-GPU-relative speedup per GPU. (c) Throughput with 25th--75th percentile error bars.}
  \label{fig:dlrm-performance}
\end{figure*}

Figure~\ref{fig:dlrm-performance} shows that SparseLift's relative gains increase with GPU count. Its single-GPU performance is comparable to DLRM-Baseline, while it achieves model-execution speedups of $1.19\times$, $1.37\times$, and $1.60\times$ on 2, 4, and 8 GPUs, respectively, increasing 8-GPU throughput from 1.837 to 2.937 M samples/s. With a fixed global batch size, computation per GPU decreases as GPU count grows, leaving less computation over which to amortize the overhead of separate communication calls and making communication aggregation increasingly beneficial. Strong-scaling efficiency on 8 GPUs improves from 19.8\% to 31.7\%, further supporting the interpretation that aggregation reduces communication startup and scheduling overhead to improve multi-GPU execution efficiency. SparseLift matches TorchRec closely on 4 GPUs and achieves a $1.26\times$ speedup on 8 GPUs, demonstrating that its automatic optimization remains competitive in the multi-GPU configurations of this public benchmark.

\begin{table}[!h]
  \centering
  \caption{DLRM quality after one epoch. Loss MAE/RMSE compare 24 matched steps with DLRM-Baseline, scaled by $10^5$.}
  \label{tab:dlrm-quality}
  \footnotesize
  \setlength{\tabcolsep}{2.5pt}
  \begin{tabular}{@{}lrrrrr@{}}
    \toprule
    System & ROC-AUC & AP & LogLoss & MAE & RMSE \\
    \midrule
    DLRM-Baseline & 0.7884 & 0.5832 & 0.46541 & --- & --- \\
    SparseLift & 0.7883 & 0.5829 & 0.46575 & 1.975 & 3.645 \\
    \bottomrule
  \end{tabular}
\end{table}

\begin{figure}[!b]
  \centering
  \includegraphics[scale=0.81]{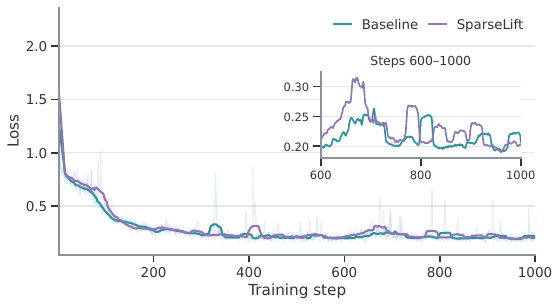}
  \caption{Industrial training loss. Faint/solid curves show per-step values/25-step centered means.}
  \label{fig:industry-loss}
\end{figure}

\subsection{Training Consistency}
Figure~\ref{fig:industry-loss} shows similar downward loss trends for the industrial Baseline and SparseLift, with mean losses of 0.21098 and 0.20926 over the final 100 steps, respectively, indicating that execution reorganization does not materially alter the observed training trajectory. As shown in Table~\ref{tab:dlrm-quality}, the mean absolute difference and root mean square difference in DLRM loss at common logged steps are $1.975\times10^{-5}$ and $3.645\times10^{-5}$, respectively, while ROC-AUC, AP, and LogLoss remain close after one training epoch. Training trajectories provide evidence of consistent learning dynamics, complemented by final metrics that assess training quality. Together, these results support comparable training behavior under SparseLift in the evaluated configurations, with no evident degradation in training quality accompanying the performance gains.

%% file: tex/4-Related.tex
\textbf{Parameter Sharding.} Parameter sharding methods address embedding-table partitioning, placement, and load balancing, using parameter sizes, access patterns, and device resources to determine how parameters are organized. AutoShard, DreamShard, and NeuroShard combine learned cost estimation with search to guide sharding and placement decisions~\cite{10.1145/3534678.3539034,Zha2022DreamShardGE,Zha2023PretrainAS}; RecShard uses access distributions to plan embedding-parameter placement across memory tiers~\cite{Sethi2022RecShardSF}. These methods primarily make decisions about parameter partitions and their mapping to hardware resources.

\textbf{Specialized Training Systems.} Specialized training systems organize communication, computation, and input processing within their respective runtimes. TorchRec provides automatic sharding and pipelines data transfer, communication, and computation~\cite{torchrec2022}; PICASSO combines packing, resource interleaving, and hot-entry caching~\cite{Zhang2022PICASSOUT}; FusedRec aggregates embedding communication~\cite{Huang2026FusedRecFE}; and EMBark jointly optimizes sharding, communication compression, and pipelining~\cite{Liu2024EmbeddingOF}. RAP targets preprocessing fusion and resource-aware co-execution~\cite{Wang2024RAPRA}, while TicTac uses communication ordering to manage iteration time and its variability in distributed training~\cite{Hashemi2018TicTacAD}. These systems focus on specific execution mechanisms and their associated scheduling relationships.

\textbf{Domain-Specific Compilers.} Compilers for recommendation workloads primarily target inference graphs or local embedding computations. RECom uses an Embedding Column representation to support cross-column fusion, while RACE optimizes inference graphs and device placement~\cite{Pan2023REComAC,10.1145/3694906.3743310}. RecFlex uses interference-aware tuning to select feature schedules and generate fused kernels~\cite{Pan2024RecFlexEF}; Ruggeri et al. study embedding-lookup dataflows and multicore mappings~\cite{10818204}; and Ember uses hierarchical intermediate representations to generate embedding code for decoupled access-execute architectures~\cite{11395192}. These approaches primarily optimize explicitly represented inference graphs or embedding computation regions, without a unified treatment of cross-stage and cross-iteration execution relations in sparse training paths.

%% file: tex/5-Conclu.tex
We present SparseLift, available at \url{https://anonymous.4open.science/r/SparseLift}, which brings runtime-internal computation, communication, and execution relations into automatic graph optimization through sparse semantic lifting, incremental cost-based decisions, and constraint-aware planning. On an industrial video-platform task with 1B dense parameters and 100B sparse parameters, SparseLift improves steady-state throughput by 16.5\%. DLRM further demonstrates workload-specific strategy selection and performance gains. These results show that recovering sparse execution relations and using them to select and organize optimizations can translate hidden opportunities into industrial training gains. Future work will validate SparseLift in additional industrial recommendation settings and explore joint compilation and optimization of sparse paths and dense computation.

%% file: tex/6-Decision-Details.tex
This appendix details the parameterization, candidate construction, and calibration procedures used by the reference implementation of the automatic decision rules in Section~\ref{sec:incremental-planning}.

\textbf{Decision-Rule Parameters.} The reference implementation uses $q=0.75$ and computes quantiles by linear interpolation between adjacent order statistics. The parameters $(\epsilon_s,\tau_s,\kappa_s)$ are $(1\,\mathrm{ms},0.25,2)$ for latency hiding and $(1\,\mathrm{ms},0.10,2)$ for code generation. Subgraph parallelism uses $(\epsilon_p,\tau_p,\kappa_v,\kappa_w)=(0,0.05,3,2)$, and communication aggregation uses $(\epsilon_a,\tau_a,\lambda_a)=(0,0.05,0.5)$. These coefficients combine with calibrated response scales and interference budgets to determine admission tolerances and comparison bands; candidates are selected automatically using recovered semantics and relative costs.

\textbf{Structural Tie-Breaking for Generated Implementations.} For supported gradient-preparation regions, we instantiate $B_{\mathcal R}$ using the workload proxy below. Here, $N$ is the member count and $d_i$ is the embedding dimension. For each aligned observation $t$, $k_{i,t}$ and $u_{i,t}$ are the respective cross-rank maxima of the member's key count and unique-row count. We then compute workload and take its median across observations. The ratio $\chi$ captures skew among members with positive workloads and is zero for an empty set. This criterion applies only to nonnegative measured increments within tolerance; admission still depends on paired measurements of the complete region.
\begin{gather*}
w_i=\operatorname{med}_t[(k_{i,t}+u_{i,t})d_i],\quad I_+=\{i:w_i>0\},\displaybreak[1]\\
\chi=\begin{cases}
0, & I_+=\emptyset,\\
\displaystyle\frac{\max_{i\in I_+}w_i}{\operatorname{med}_{i\in I_+}w_i}, & \text{otherwise},
\end{cases}\displaybreak[1]\\
a_N=\min\{1,(N-1)/4\},\quad b_\chi=\min\{1,(\chi-1)_+\},\displaybreak[1]\\
z_{\mathrm{ref}}=\tfrac12a_N(1+b_\chi),\displaybreak[1]\\
B_{\mathcal R}=[z_{\mathrm{ref}}\ge0.75]\land[\chi\ge1.5].
\end{gather*}

\textbf{Candidate Construction.} The reference industrial executor exposes group counts of $\{1,2\}$, $\{1,2,4\}$, and $\{1,4,6\}$ for input preparation, output assembly, and gradient merging, respectively, excluding counts greater than the number of members. Each $\pi_G$ partitions the fixed member order into contiguous groups whose sizes differ by at most one member, with larger groups first. For gradient preparation, the decision compares nongenerated and generated implementations. When the generated implementation is selected, the planner binds it to one group as required by its contract. The executor-provided candidate sets delimit the available choices, from which the decision algorithm automatically selects execution configurations based on region-level incremental costs.

\textbf{Bounded Calibration.} Table~\ref{tab:reference-calibration} gives the execution budget for one calibration attempt at each measurement point in the reference implementation. Parallelism and code generation interpolate neighboring reference anchors to construct paired increments, while communication pairs alternate execution order. Enabled latency-hiding observations count only when the intended schedule executes and its prefetched results are consumed. These budgets apply per measurement point; feasible candidates and their query requirements determine the calibration scope. The selected plan is reused during subsequent steady-state training of the same task.

\begin{table}[tb]
  \centering
  \caption{Reference calibration budgets per measurement point and attempt. Warmup is excluded from observations.}
  \label{tab:reference-calibration}
  \small
  \setlength{\tabcolsep}{4pt}
  \begin{tabularx}{\columnwidth}{@{}>{\raggedright\arraybackslash}p{0.28\columnwidth}>{\raggedright\arraybackslash}X@{}}
    \toprule
    Strategy & Warmup and measured executions \\
    \midrule
    Communication aggregation & One warmup pair, then two measured pairs per compatible role or complete forward chain. \\
    Subgraph parallelism & One warmup per variant, then a bidirectional sweep with two region observations per group count. \\
    Code generation & One warmup per implementation, then complete-region measurements in the order $\mathrm{ref}\to\mathrm{gen}\to\mathrm{gen}\to\mathrm{ref}$. \\
    Latency hiding & One warmup step, then two valid complete-step observations per schedule. \\
    \bottomrule
  \end{tabularx}
\end{table}

The reference implementation obtains the host scheduling time scale $H^{cal}$ from a warmed worker fan-out/fan-in probe. Each rank submits four empty tasks to a four-worker pool and waits for completion over nine rounds. We discard the first two rounds, take the median duration of the remaining rounds, and then take the cross-rank $q$-quantile. This time scale informs orchestration tolerance, while region measurements capture workload and contention during actual execution.

In Algorithm~\ref{alg:sequential-planning}, \texttt{BuildEvidence} and \texttt{SelectCandidate} implement evidence construction and the four selection rules in the main text. \texttt{MergeBindings} implements $\oplus_{\mathcal C}$ under the transformation contracts while preserving other region bindings, and \texttt{FreezePlan} fixes the plan for subsequent execution.

\section{Industrial Integration and Design Rationale}
\label{app:industrial-integration}

This appendix expands the industrial integration described in Section~\ref{sec:implementation}, focusing on how the existing parameter servers and training pipeline shape the compilation interface and execution organization.

\textbf{Preserving Production Runtime Boundaries.} We integrate graph optimization through separate model and training-loop entry points. Before the first actual forward pass, the model entry point locates and compiles sparse modules in place. Instance-scoped lifecycle wrappers expose complete optimizer steps, input readiness, and result consumption to the compiler and are restored when the training invocation exits. This integration preserves the platform's training-loop implementation and global trainer types. Existing parameter servers continue to manage sparse-parameter storage and access, while native iterator call sites and dense and sparse optimizer steps retain their original behavior. The compiler can therefore reorganize sparse execution within the platform's established parameter-management and training mechanisms.

\textbf{Making Runtime Facts Checkable.} The adapter exports table and operation identities, typed ports, communication layouts, parameter versions, and workspace ownership, from which the compiler constructs regions. For example, communication cost queries use per-member source--destination traffic in bytes, while execution contracts retain data types and element-count splits and check receive splits against payload sizes. Keeping these units explicit gives cost comparisons and communication execution a consistent semantic basis. Missing or ambiguous facts remain unknown and prevent selection of candidates that require them. Applying the method to another runtime requires adapting fact export, lifecycle integration, and candidate execution contracts, followed by calibration in the target environment.

\textbf{Bounding Early Execution by Lifecycle Visibility.} The industrial platform migrates resources at sparse-cache period transitions, but its interface exposes only the return of the native iterator, without complete readiness dependencies for internal cache streams. We therefore retain the formal forward path for the first batch of a period and do not submit its sparse preparation or associated communication ahead of the iteration. Period identity comes from the actual prefetcher. This design reflects that parameter-independent computation can still depend on cache and workspace lifetimes. Preserving the original path at boundaries with incomplete readiness evidence allows subsequent ordinary batches to use validated overlap windows without assuming additional platform-internal behavior.

\textbf{Distributed Activation and Submission Boundaries.} To preserve collective ordering under schedule changes, we create and warm up dedicated communication groups at boundaries traversed by all ranks in the same order; subsequent data paths only read ready handles. Sparse-gradient communication uses a separate group from DDP's default group. Activation updates only configuration fields explicitly bound by the new plan, preserving prior choices elsewhere. If plan checks fail before submission, ranks jointly retain the validated plan. Communication-group initialization failure cannot instead be handled by individual ranks switching to the default group. If execution fails after partial submission, the system terminates execution after synchronization and resource cleanup to prevent duplicate communication or writes. This distinction separates a rejectable plan choice from an execution failure that cannot safely be replayed.

\section{Detailed Industrial Benchmark Results}
\label{app:industrial-benchmarks}

This appendix reports the numerical results and decision evidence underlying the industrial figures in the main text. All experiments use the same video-platform recommendation task, with 1 or 8 GPUs on one node and 16 GPUs across two nodes; dedicated parameter servers store the sparse parameters. Within each configuration, the comparison preserves native parameter access and dense-path optimizations and evaluates native sparse execution against SparseLift's automatic plan.

\subsection{Throughput and Stage Accounting}

Table~\ref{tab:industrial-throughput-details} reports the throughput distribution on 8 GPUs. Each system contributes 94 periods from the first 1000 steps after excluding the first 60 steps and the transition step at the start of each period. Period throughput divides the total sample count of the nine ordinary steps by their total execution time, with profiling enabled. The quartiles describe the distribution across periods, and median throughput improves by 16.5\%.

\begin{table}[tb]
  \centering
  \caption{Industrial ordinary-step throughput on 8 GPUs, in samples/s, over 94 steady-state periods per system.}
  \label{tab:industrial-throughput-details}
  \small
  \setlength{\tabcolsep}{4pt}
  \begin{tabular}{@{}lrrr@{}}
    \toprule
    System & 25th pct. & Median & 75th pct. \\
    \midrule
    Baseline & 4260.4 & 4334.6 & 4370.3 \\
    SparseLift & 4982.8 & 5050.7 & 5095.8 \\
    \bottomrule
  \end{tabular}
\end{table}

\textbf{Stage Measurements.} Table~\ref{tab:industrial-stage-details} expands the complete stage accounting in Figure~\ref{fig:industry-ablation}(a). From five recorded profiler steps per configuration, we exclude one observation marked as a transient spike in the measurement record and compute stage means and the sample standard deviation of total step time over the four retained observations. Sparse forward and backward include communication. Other forward and backward times subtract the sparse components from their respective totals, and Other subtracts complete forward and backward times from total step time.

\begin{table*}[tb]
  \centering
  \caption{Industrial stage timings in ms over four retained observations per row. SD is the sample standard deviation of total step time. Component and total means are rounded independently.}
  \label{tab:industrial-stage-details}
  \small
  \setlength{\tabcolsep}{4pt}
  \begin{tabular}{@{}clrrrrrrr@{}}
    \toprule
    GPUs & System & \shortstack{Sparse\\forward} & \shortstack{Other\\forward} & \shortstack{Sparse\\backward} & \shortstack{Other\\backward} & Other & \shortstack{Total\\step} & SD \\
    \midrule
    1 & Baseline & 26.75 & 112.50 & 25.50 & 213.25 & 50.75 & 428.75 & 0.50 \\
    1 & SparseLift & 2.50 & 110.75 & 9.75 & 214.25 & 32.50 & 369.75 & 1.50 \\
    \midrule
    8 & Baseline & 44.25 & 113.75 & 28.00 & 222.75 & 42.75 & 451.50 & 7.14 \\
    8 & SparseLift & 3.13 & 108.88 & 9.13 & 227.88 & 29.50 & 378.50 & 2.38 \\
    \midrule
    16 & Baseline & 48.00 & 110.50 & 39.50 & 229.00 & 47.25 & 474.25 & 3.40 \\
    16 & SparseLift & 6.00 & 105.75 & 23.50 & 233.25 & 33.50 & 402.00 & 3.56 \\
    \bottomrule
  \end{tabular}
\end{table*}

On 8 GPUs, 60.0 ms of the 73.0 ms difference in sampled step time lies in the exposed sparse-forward and sparse-backward components. The forward reduction is consistent with reduced consumer waiting when preparation executes early, while the backward difference is consistent with reorganizing compatible communication and gradient computation across tables. Total step time decreases by 13.8\%, 16.2\%, and 15.2\% on 1, 8, and 16 GPUs, respectively, across both single-node and two-node configurations of the same task.

\subsection{Observed Decisions and Strategy Contributions}

\textbf{Complete-Schedule Decisions.} Table~\ref{tab:industrial-lookahead-calibration} reports rank-level evidence for the latency-hiding decision on 8 GPUs. The reference already includes region implementations selected by preceding decisions, so calibration compares the incumbent plan with its one-step-lookahead candidate. Each rank performs one warmup and retains two valid complete-step measurements per schedule, requiring actual consumption of prefetched results on the enabled path. The candidate minimum is lower on all eight ranks. Taking the cross-rank 75th percentile separately for each schedule reduces the response from 418.752 to 384.131 ms. This evidence supports admitting latency hiding under the main-text rule, and execution records show identical plan bindings across the eight ranks.

\begin{table}[tb]
  \centering
  \caption{Industrial lookahead calibration in ms. Each rank reports its minimum of two measurements; the aggregate increment is $F(T_1)-F(T_0)$.}
  \label{tab:industrial-lookahead-calibration}
  \small
  \setlength{\tabcolsep}{3pt}
  \begin{tabular}{@{}lrrr@{}}
    \toprule
    Rank & Incumbent & Lookahead & Increment \\
    \midrule
    0 & 418.464 & 383.889 & $-34.575$ \\
    1 & 418.752 & 384.108 & $-34.644$ \\
    2 & 418.752 & 383.996 & $-34.756$ \\
    3 & 418.684 & 384.095 & $-34.589$ \\
    4 & 418.953 & 383.948 & $-35.005$ \\
    5 & 418.676 & 384.198 & $-34.479$ \\
    6 & 418.477 & 384.364 & $-34.114$ \\
    7 & 418.731 & 383.998 & $-34.733$ \\
    \midrule
    Aggregate & 418.752 & 384.131 & $-34.621$ \\
    \bottomrule
  \end{tabular}
\end{table}

\textbf{Contributions within the Combined Plan.} Table~\ref{tab:industrial-ablation-details} reports absolute latency and relative changes after strategy removal. Ablation follows the exclusions marked in its measurement records, retaining four observations per configuration and three when latency hiding is disabled; its means are computed separately from the stage-breakdown table. Full has the lowest recorded mean latency among these configurations. Removing latency hiding increases that mean by 48 ms; the other removals increase it by 4--11 ms. Calibration compares a candidate with the incumbent, ablation measures conditional contributions within Full, and throughput evaluates the final plan against the production baseline. Together, these measurements connect decision evidence to industrial performance at distinct levels.

\begin{table}[tb]
  \centering
  \caption{Industrial strategy ablation on 8 GPUs. $n$ counts retained step observations; increases are relative to Full and are not additive.}
  \label{tab:industrial-ablation-details}
  \small
  \setlength{\tabcolsep}{3pt}
  \begin{tabular}{@{}lrrr@{}}
    \toprule
    Configuration & $n$ & Step (ms) & Increase \\
    \midrule
    Baseline & 4 & 451 & 19.3\% \\
    Full & 4 & 378 & --- \\
    Without aggregation & 4 & 383 & 1.3\% \\
    Without parallelism & 4 & 382 & 1.1\% \\
    Without latency hiding & 3 & 426 & 12.7\% \\
    Without code generation & 4 & 389 & 2.9\% \\
    \bottomrule
  \end{tabular}
\end{table}

\section{Public-Benchmark Configuration and Measurement Details}
\label{app:evaluation-details}

This appendix supplements Section~\ref{Eval} with public-benchmark configuration and statistical details, complementing the industrial evaluation. Hardware and software versions are listed in Table~\ref{tab:experimental-setup}.

\textbf{DLRM Configuration.} Table~\ref{tab:dlrm-configuration-details} specifies additional architectural settings and the random seed. All three implementations use a global batch size of 16384 and an embedding dimension of 256. TorchRec specifies batch size per rank, so its setting is $16384/R$ on $R$ GPUs. It uses \texttt{DistributedModelParallel} and \texttt{TrainPipelineSparseDist}, with embedding sharding selected by its native planner and embedding computation executed through FBGEMM. DLRM-Baseline and SparseLift share preprocessing and data splits, while TorchRec retains its native preprocessing, categorical hashing, and data splits. Accordingly, the training-quality table uses the same-stack comparison, and TorchRec serves as a cross-stack performance reference.

\begin{table}[tb]
  \centering
  \caption{Additional DLRM settings. MLP widths specify layer outputs, excluding input dimensions.}
  \label{tab:dlrm-configuration-details}
  \small
  \setlength{\tabcolsep}{4pt}
  \begin{tabularx}{\columnwidth}{@{}>{\raggedright\arraybackslash}p{0.48\columnwidth}>{\raggedright\arraybackslash}X@{}}
    \toprule
    Setting & Value \\
    \midrule
    Dense input features & 13 \\
    Bottom MLP widths & 512--256--64--256 \\
    Top MLP widths & 512--256--1 \\
    Feature interaction & Pairwise dot products \\
    Random seed & 123 \\
    \bottomrule
  \end{tabularx}
\end{table}

\textbf{DLRM Timing and Aggregation.} Each run uses 50 warmup steps followed by 450 measured steps in nine consecutive 50-step windows. At each window boundary, we synchronize, sum the recorded per-step CUDA-event durations on each rank, take the maximum sum across ranks, and divide by 50 to obtain mean step time. Timing covers forward execution, communication, loss computation, backward execution, and the optimizer, excluding data loading and host-to-GPU transfers. Window throughput is the global batch size divided by this mean step time in seconds. We take the median over nine windows per run and then the median of the five independent run medians. Throughput error bars give the 25th--75th percentile interval over these five values. All measured windows are retained in the reported model-execution statistics.